\documentclass[10pt,twocolumn,letterpaper]{article}

\usepackage[pagenumbers]{cvpr} 

\usepackage{graphicx}
\usepackage{amsmath}
\usepackage{amssymb}
\usepackage{booktabs}

\usepackage{multirow} 
\usepackage{amsfonts}
\usepackage{pifont}
\newcommand{\cmark}{\ding{51}}%
\newcommand{\xmark}{\ding{55}}%
\usepackage{makecell}
\usepackage{soul}
\usepackage[accsupp]{axessibility}

\usepackage[pagebackref,breaklinks,colorlinks]{hyperref}

\usepackage[capitalize]{cleveref}
\crefname{section}{Sec.}{Secs.}
\Crefname{section}{Section}{Sections}
\Crefname{table}{Table}{Tables}
\crefname{table}{Tab.}{Tabs.}

\newcommand{\siou}{m\_sIoU} 

\def\cvprPaperID{1519} 
\def\confName{CVPR}
\def\confYear{2022}

\begin{document}

\title{TubeDETR: Spatio-Temporal Video Grounding with Transformers}

\author{Antoine Yang$^{1,2}$, Antoine Miech$^{3}$, Josef Sivic$^{4}$, Ivan Laptev$^{1,2}$, Cordelia Schmid$^{1,2}$
\smallskip\\ 
\small{$^1$Inria Paris \quad $^2$D\'{e}partement d'informatique de l'ENS, CNRS, PSL Research University \quad $^3$DeepMind \quad $^4$CIIRC CTU Prague}
\\
\small{\url{https://antoyang.github.io/tubedetr.html}} \\
\small{Note that vIoU results are updated compared to the CVPR'22 camera-ready version.}
}
\maketitle

\begin{abstract}

We consider the problem of localizing a spatio-temporal tube in a video corresponding to a given text query. 
This is a challenging task that requires the joint and efficient modeling of temporal, spatial and multi-modal interactions.
To address this task, we propose TubeDETR, a transformer-based architecture inspired by the recent success of such models for text-conditioned object detection. 
Our model notably includes:
(i) an efficient video and text encoder that models spatial multi-modal interactions over sparsely sampled frames and
(ii) a space-time decoder that jointly performs spatio-temporal localization.
We demonstrate the advantage of our proposed components through an extensive ablation study.
We also evaluate our full approach on the spatio-temporal video grounding task and demonstrate improvements over the state of the art on the challenging VidSTG and HC-STVG benchmarks.
\end{abstract}

\footnotetext[4]{Czech Institute of Informatics, Robotics and Cybernetics at the Czech Technical University in Prague.}

\vspace{-0.5cm}
\section{Introduction}\label{sec:intro}
\begin{figure}[t]
\centering
\includegraphics[width=1.\linewidth]{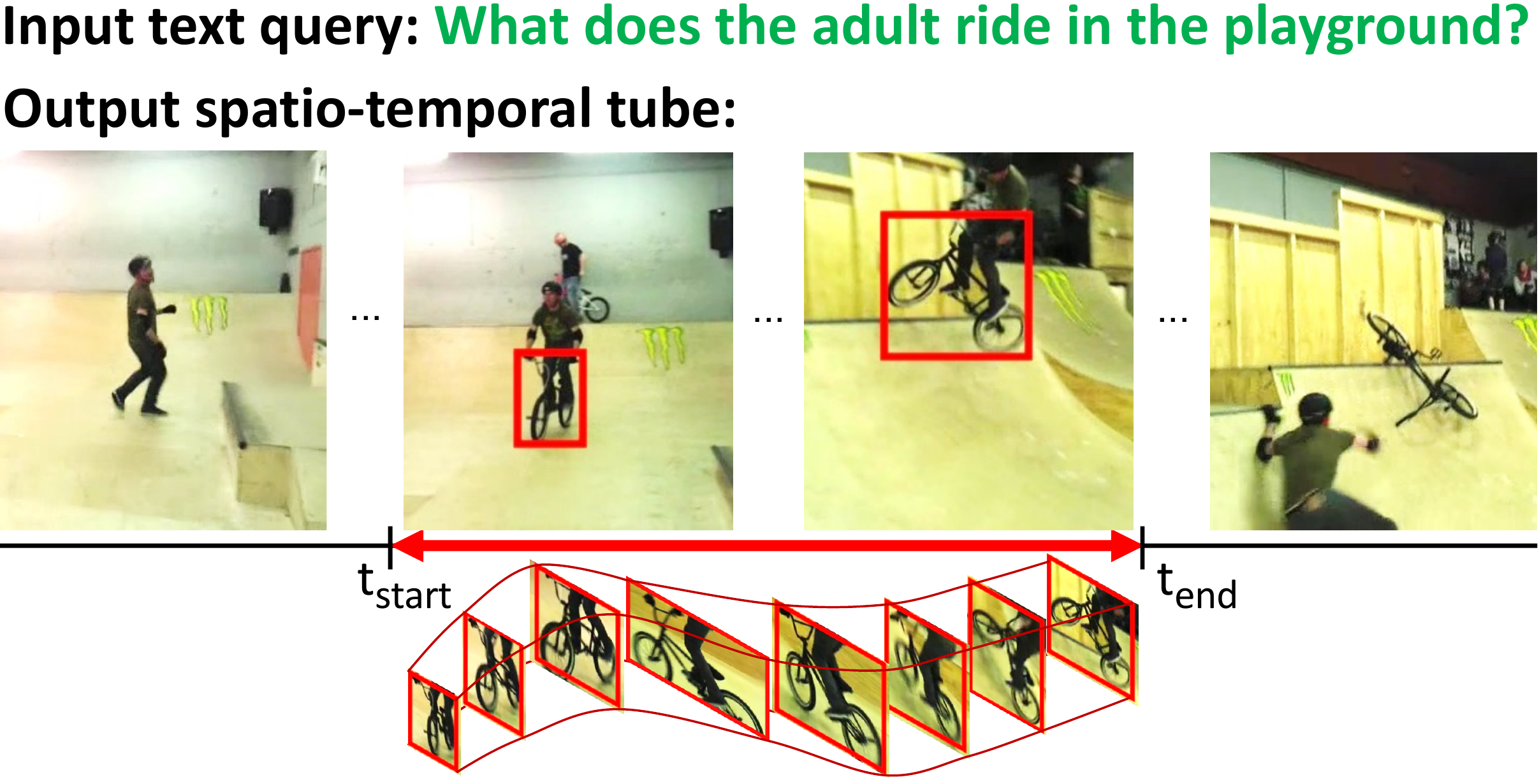}
\vspace{-0.8cm}
\caption{\small Spatio-temporal video grounding requires reasoning about space, time, and language.}
\label{fig:teaser}
\vspace{-0.7cm}
\end{figure}

Grounding natural language in visual content is a fundamental skill to build powerful and explainable vision and language models. 
In particular, understanding the association of language with spatial regions and temporal boundaries in videos is particularly important to analyze and improve multi-modal video models.
This goes beyond associating a global visual representation with a textual representation~\cite{radford2021learning, miech20endtoend}, as it requires to reason about detailed spatio-temporal visual representations and their association with natural language, as illustrated in Figure~\ref{fig:teaser}.

Spatio-temporal video grounding, recently introduced in~\cite{vidstg}, is an interesting and challenging task that lies at the intersection of visual grounding~\cite{hu2016natural, nagaraja2016modeling, vasudevan2018object} and temporal localization~\cite{hendricks17localizing, gao2017tall, chen2018temporally}. 
Given an untrimmed video and a textual description of an object, spatio-temporal video grounding aims at localizing a spatio-temporal tube (\ie, a sequence of bounding boxes) for the target object described by the input text.
This task is particularly challenging as videos are highly diverse and often present challenging scenarios where different entities have similar appearance or perform similar actions within one scene.

The success of attention-based models in natural language processing~\cite{vaswani2017attention, bert18} has recently inspired approaches to integrate transformers into computer vision tasks, such as image classification~\cite{dosovitskiy2021an}, object detection~\cite{carion2020end}, semantic segmentation~\cite{liu2021swin} or action recognition~\cite{arnab2021vivit, bertasius2021space, zhang2021vidtr, patrick2021keeping}.
Notably, with DETR~\cite{carion2020end}, transformers have shown competitive performance on object detection while removing the need of multiple hand-designed components encoding a prior knowledge about this task.
More recently, MDETR~\cite{mdetr} has extended this framework for various text-conditioned object detection tasks in the image domain, such as phrase grounding, referring expression comprehension and segmentation.

Inspired by these works, and the fact that attention-based architectures are an intuitive choice for modelling multi-modal and spatio-temporal contextual relationships in videos, we develop a transformer encoder-decoder model for spatio-temporal video grounding, as illustrated in Figure~\ref{fig:overview}.
While existing approaches for this task rely on pre-extracted object proposals~\cite{vidstg}, tube proposals~\cite{hcstvg} or upsampling layers~\cite{stvgbert}, our architecture simply reasons about abstractions called \textit{time queries} to jointly perform temporal localization and visual grounding. 
Our framework enables to use the same representations for both subtasks in order to learn powerful contextualized representations.

More specifically, our architecture includes key components to jointly model temporal, spatial and multi-modal interactions.
Our video-text encoder efficiently encodes spatial and multi-modal interactions by computing these interactions over sparsely sampled frames, and separately recovers temporally local information with a lightweight fast branch. 
Our space-time decoder models temporal interactions with temporal self-attention layers, and spatial and multi-modal interactions with time-aligned cross-attention layers.
Spatio-temporal video grounding is then tackled with multiple heads on top of the decoder outputs, which predict the object boxes and temporal start and end probabilities.
We conduct various ablation studies, where we notably show the benefit of our video-text encoder in terms of performance-memory trade-off, and the efficiency of our space-time decoder in terms of spatio-temporal grounding results. 
Finally, we show that our method significantly improves over state-of-the-art methods on two benchmarks, VidSTG~\cite{vidstg} and HC-STVG~\cite{hcstvg}. 

\noindent In summary, our contributions are three-fold:
\textit{(i)} We propose a novel architecture for spatio-temporal video grounding that performs this task with a space-time transformer decoder.
\textit{(ii)} We propose a dual-stream encoder that efficiently encodes spatial and multi-modal interactions, based on a slow multi-modal stream and a lightweight fast visual stream. \textit{(iii)} We conduct comprehensive experiments on two benchmarks, VidSTG and HC-STVG, showing the effectiveness of our framework for the spatio-temporal video grounding task. Our approach, referred to as TubeDETR, outperforms all state-of-the-art methods by a large margin.\\
Code and trained models are publicly available at~\cite{tubedetrwebpage}.

\section{Related Work}\label{sec:background}
\noindent \textbf{Spatio-temporal video grounding.} Visual grounding consists in spatially localizing an object given a referring expression, and has been an active area of research both in the image domain~\cite{deng2018visual, hu2016natural, hu2017modeling, nagaraja2016modeling, xiao2017weakly, yu2017joint, zhang2018grounding, zhuang2018parallel, liu2021relation, wang2021improving} and the video domain~\cite{huang2018finding, shi2019not, vasudevan2018object}.
A standard paradigm consists in using pre-extracted object proposals~\cite{liu2017referring, liu2019improving, wang2019neighbourhood, yamaguchi2017spatio, yang2019cross, yang2019dynamic, yu2018mattnet}, while some recent works~\cite{liao2020real, mdetr, deng2021transvg, huang2021look, luo2020multi, yang2019fast, yang2020improving} have proposed one-stage approaches which do not rely on such proposals.
Our work follows the one-stage framework of MDETR~\cite{mdetr}, but extends it to spatio-temporal video grounding with temporal localization losses (see Equation~\ref{eq:objective}), slow-fast encoding (see Figure~\ref{fig:encoder}), and space-time decoding (see Figure~\ref{fig:decoder}).

A separate line of work focuses on temporally localizing moments in a video given a natural language query~\cite{chen2018temporally, chen2019localizing, hendricks17localizing, hendricks2018localizing, gao2017tall, lin2020weakly, mithun2019weakly, rodriguez2020proposal, wang2019language, zhang2019cross, zhang2019man, zhang2020learning, zhang2020span,  zeng2020dense, wang2020temporally, yuan2019find, chen2019semantic, he2019read}. These works build architectures that reason about time but do not preserve spatial information.
Spatio-temporal video grounding lies at the intersection of temporal localization and visual grounding.
While some approaches~\cite{chen2019weakly, hcstvg, yamaguchi2017spatio} rely on pre-extracted tube proposals, or object proposals~\cite{vidstg}, our method does not require any pre-extracted proposals.
A recent work~\cite{stvgbert} proposes STVGBert, a one-stage approach that extends the VilBERT model~\cite{lu2019vilbert} pretrained on Conceptual Captions~\cite{sharma2018conceptual} to this task. 
STVGBert uses deconvolutions to perform visual grounding, and symmetrically models temporal and spatial interactions.
In contrast, our architecture performs visual grounding with a transformer decoder, and separately reasons about the temporal and spatial dimensions.

\noindent \textbf{Temporal modeling for video understanding.} The rise of powerful models for image understanding such as ViT~\cite{dosovitskiy2021an} or DETR~\cite{carion2020end} has fostered research  extending these models to the video domain~\cite{arnab2021vivit, bertasius2021space, zhang2021vidtr, lei2021qvhighlights, he2021end, patrick2021keeping}. 
In particular, Lei \etal~\cite{lei2021qvhighlights} propose an architecture that views moment retrieval as a direct set prediction problem, but is unsuitable to visual grounding as it does not preserve spatial information.
He \etal~\cite{he2021end} extend the DETR framework to videos, and propose an architecture built with sequentially added modules on top of Deformable DETR~\cite{zhu2020deformable}, while ours is built on inner modifications of a pretrained encoder and decoder and also reasons about language.
Our dual-branch encoder is also related to SlowFast networks~\cite{feichtenhofer2019slowfast, xiao2020audiovisual} which combine fast and slow video streams. 
In contrast, in our case, both streams operate on features extracted from the same backbone, and our dual-stream architecture is motivated by the computational complexity related to multi-modal modeling.

\noindent \textbf{Vision and language.} Transformer-based architectures have become ubiquitous in various vision and language tasks~\cite{su2019vl, tan2019lxmert, lu2019vilbert, chen2019uniter, desai2020virtex, huang2020pixel, li2019unicodervl, li2020oscar, lu202012, zhou2020unified, chen2021history, kim2021vilt, cornia2020meshed}. 
Most video-text transformers rely either on pre-extracted object features~\cite{zhu2020actbert}, or spatially pooled features~\cite{sun2019videobert, li2020hero, ging2020coot, yang2021just, gabeur2020multi, zhou2018endtoend}, which do not preserve detailed spatial information. 
In contrast, our architecture is designed to preserve spatial information to perform visual grounding.
Some recent works propose transformer-based architectures reasoning on videos and text that do preserve spatial information~\cite{lei2021less, zellersluhessel2021merlot, akbari2021vatt, bain2021frozen}. 
However, these works typically aim to learn global video representations to tackle video-level prediction tasks, while we focus on learning detailed frame-level representations to address a dense prediction task requiring spatial and temporal localization.

\section{Method}\label{sec:method}
\begin{figure*}[t]
\centering
\includegraphics[width=1.\linewidth]{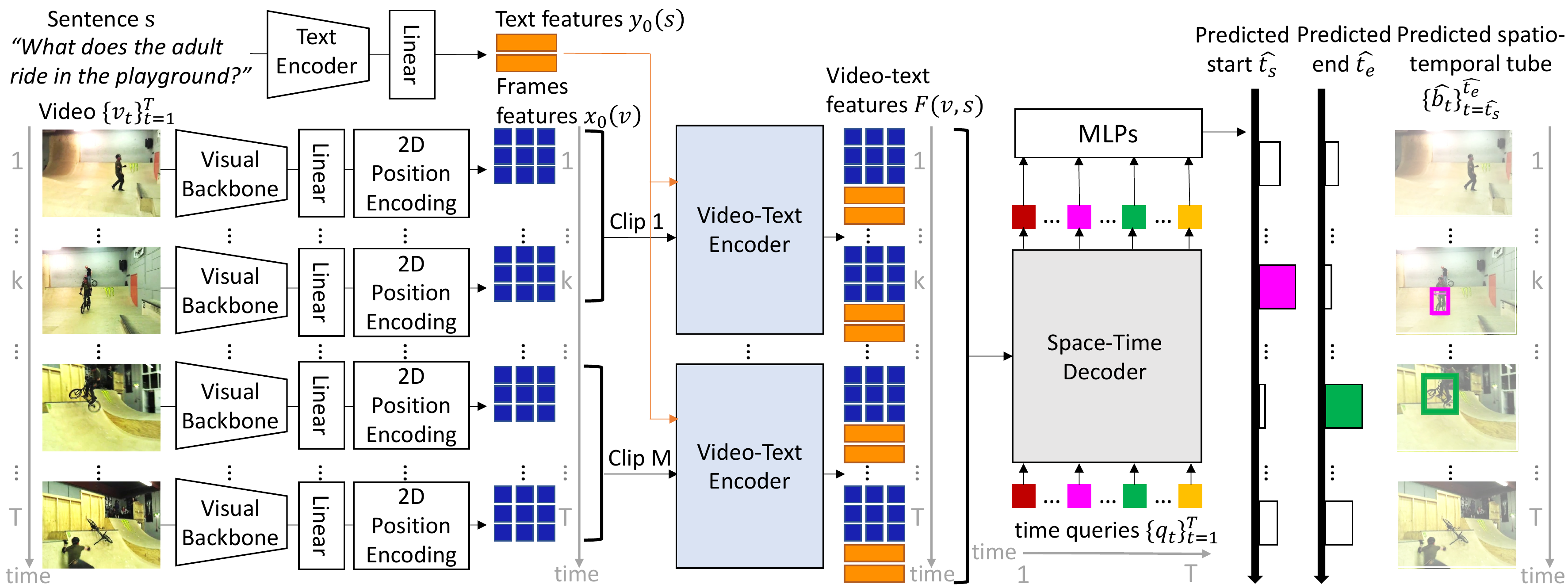}
\vspace{-0.7cm}
\caption{\small \textbf{TubeDETR model overview}. 
All input video frames $v_t$ and the sentence $s$ are first processed with a Visual Backbone and a Text Encoder.
The resulting text and video features $y_0(s)$ and $x_0(v)$ are then jointly encoded with a Video-Text Encoder that computes spatial and multi-modal interactions for $M$ short clips of $k$ frames (about 1 second).
The resulting video-text features $F(v,s)$ are then decoded into the output spatio-temporal tube $\hat{b}$ using a Space-Time Decoder that jointly reasons about time, space and text over the entire video. 
}
\vspace{-0.5cm}
\label{fig:overview}
\end{figure*}

We first give an overview of our model in Section~\ref{sec:overview}. 
Next, we describe in detail the two main components of our model, the video-text encoder (Section~\ref{sec:encoder}) and the space-time decoder (Section~\ref{sec:decoder}).
Then in Section~\ref{sec:losses} we explain the loss used to train our model. 
Finally in Section~\ref{sec:weights} we present how we initialize our model weights.

\subsection{Overview}\label{sec:overview}
Our objective is, given a video and a language query, to output a spatio-temporal tube, \ie a sequence of bounding boxes with temporal boundaries, grounding the language query in the video. 
This is challenging as it requires modelling long-range {\em spatial} and {\em temporal} interactions between the language query and the video where the video may have hundreds of frames represented by tens of thousands spatio-temporal video features. 
Hence efficiency is a major challenge. 
To address this issue we design an encoder-decoder architecture, illustrated in Figure~\ref{fig:overview}, that enables accurate yet efficient modelling of video-language spatial and temporal interactions across the entire video.
In particular, our two-stream video-text encoder (Section~\ref{sec:encoder}) models video-language interactions only over short clips of about one second but allows for detailed spatial localization. Our space-time decoder (Section~\ref{sec:decoder}) then models long-range temporal interaction over the entire video to produce a temporally consistent output and accurate predictions of the start and end times of the output spatio-temporal tube. 

\begin{figure*}[t]
\centering
\includegraphics[width=1\linewidth]{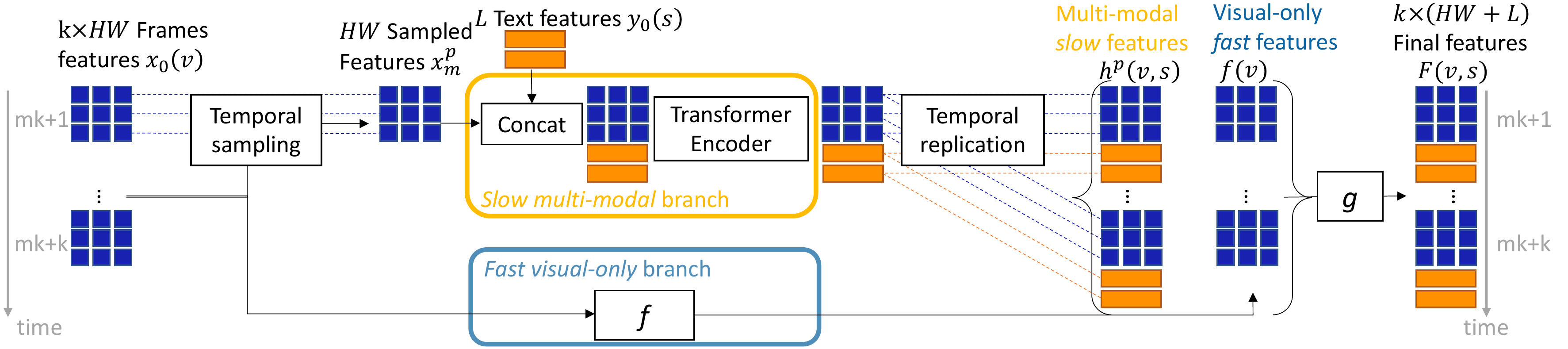}
\vspace{-0.7cm}
\caption{\small \textbf{Video-Text Encoder} takes as input a set of 2D flattened image features $x_0(v)$ together with a set of text features $y_0(s)$ from the query sentence, and outputs a set of video-text features $F(v, s)$, one for each frame.
Top: the \textit{Slow multi-modal} branch first samples video features $x^p_m$, one from every $k$ frames.
Then it computes multi-modal interactions between the sampled features $x^p_m$ and text features $y_0$ using a transformer encoder. 
The temporal sampling reduces the number of video features in order to efficiently compute the attention-based interactions. 
Bottom: lightweight ``Fast visual-only" branch $f$ processes features from {\em all} frames but without any attention layers for increased efficiency.
Features from both branches are then combined in module $g$ into the final set of per-frame features $F(v,s)$.}
\label{fig:encoder}
\vspace{-0.5cm}
\end{figure*}

\subsection{Video-Text Encoder}\label{sec:encoder}
Our encoder is illustrated in Figure~\ref{fig:encoder} and described next.
Its objective is to model spatial and multi-modal interactions between the language query and the video to accurately spatially ground the query in each frame. 
To achieve this, we leverage the ability of the self-attention layers to jointly model spatial and visual-linguistic interactions~\cite{mdetr, lei2021less, huang2020pixel}. 
However, computing self-attention between visual features and textual features for every frame is computationally expensive.
For this reason, we propose to compute spatial and multi-modal interactions only for every $k$-th frame. 
We denote the resulting stream as \textit{slow multi-modal} branch.
We use a separate lightweight \textit{fast visual-only} branch that preserves the original frame rate and allows us to recover some of the high frequency spatio-temporal details lost by the sparse sampling in the slow branch.

Formally, our encoder takes as input a set of 2D flattened image features $x_0(v) \in \mathbb{R}^{T \times HW \times d}$ from the visual backbone for all $T$ frames of the input video together with a set of $L$ text features $y_0(s) \in \mathbb{R}^{L \times d}$ extracted by the text encoder from the query sentence, and outputs a set of video-text features $F(v, s) \in \mathbb{R}^{T \times (HW+L) \times d}$, one for each frame.
Next we give the details of the Slow and Fast branches, and the final feature aggregation module.  

\noindent \textbf{Slow multi-modal branch.}
The goal of this branch (see top of Figure~\ref{fig:encoder}) is to model interactions between visual and textual representations. 
This branch first samples features from \emph{one} frame for a short clip of $k$ consecutive frames.
A typical clip length is one second, \ie $k=5$ with a standard frame rate of 5 frames per second~\cite{vidstg}. 
Formally, the resulting feature map is written as $x^p \in \mathbb{R}^{M \times HW \times d}$ where $M = \lceil \frac{T}{k} \rceil $ is the number of clips, $k$ is the length of the clip and $T$ is the length of the entire video. 
We then concatenate, for each clip $m$, its visual features $x^p_m$ with text features $y_0(s)$ and forward it to a N-layer transformer encoder.
The outputs are contextualized visual-text representations $h^p(v,s) \in \mathbb{R}^{M \times (HW + L) \times d}$, which effectively combine information from the input video $v$ and the query sentence $s$.

\noindent \textbf{Fast visual-only branch.} 
The previously explained temporal sparse sampling scheme reduces significantly the memory requirements of the video-text encoder but results in a loss of spatio-temporal details which are important for spatio-temporal video grounding.
To alleviate this issue, we introduce module $f$ (see bottom of Figure~\ref{fig:encoder}) which operates on \emph{2D flattened image features for all frames}.
Formally, given feature map $x_0(v)$, this module outputs visual features $f(v) \in \mathbb{R}^{T \times HW \times d}$.
This \textit{fast} branch preserves the spatial and temporal resolution of the features but is computationally light as it does not compute any multi-modal or spatial interactions. 
For additional efficiency, at training time, this branch does not back-propagate gradients to the visual backbone.
Furthermore, we show in Section~\ref{sec:ablations} that it is able, when combined with the temporally sparse features obtained from the slow branch, to recover some of the temporal information lost during the temporal sampling.

\noindent \textbf{Slow-Fast feature aggregation.} 
We now describe the \textit{slow} and \textit{fast} branches aggregation module (see Figure~\ref{fig:encoder}, right), which fuses information from both branches and outputs final video-text features.
To match the temporal dimension of the output from the \textit{fast} branch $f(v)$, the output of the \textit{slow} multi-modal branch $h^p(v,s)$ is temporally replicated $k$ times for each clip resulting in video-text encodings $h(v, s) \in \mathbb{R}^{T \times (HW+L) \times d}$.
These encodings are a concatenation of text-contextualized visual encodings $h_v(v, s) \in \mathbb{R}^{T \times HW \times d}$ and visually-contextualized textual encodings $h_s(v, s) \in \mathbb{R}^{T \times L \times d}$. 
The text-contextualized visual encodings $h_v(v, s)$ are combined with the outputs of the \textit{fast} branch with an additional aggregation module $g$ and a residual connection, resulting in aggregated visual encodings $F_v(v, s) = g(h_v(v, s), f(v)) + h_v(v, s)$. 
The final output of our video-text encoder is obtained by concatenating these aggregated visual encodings with the visually-contextualized textual encodings \ie $F(v, s) = [F_v(v, s), h_s(v, s)] \in \mathbb{R}^{T \times (HW+L) \times d}$.
In detail, the module $g$ is implemented as a sum followed by a linear layer, \ie $g(h_v(v, s), f(v)) = \texttt{Linear}(h_v(v, s) + f(v))$.

\begin{figure*}[t]
\centering
\includegraphics[width=1\linewidth]{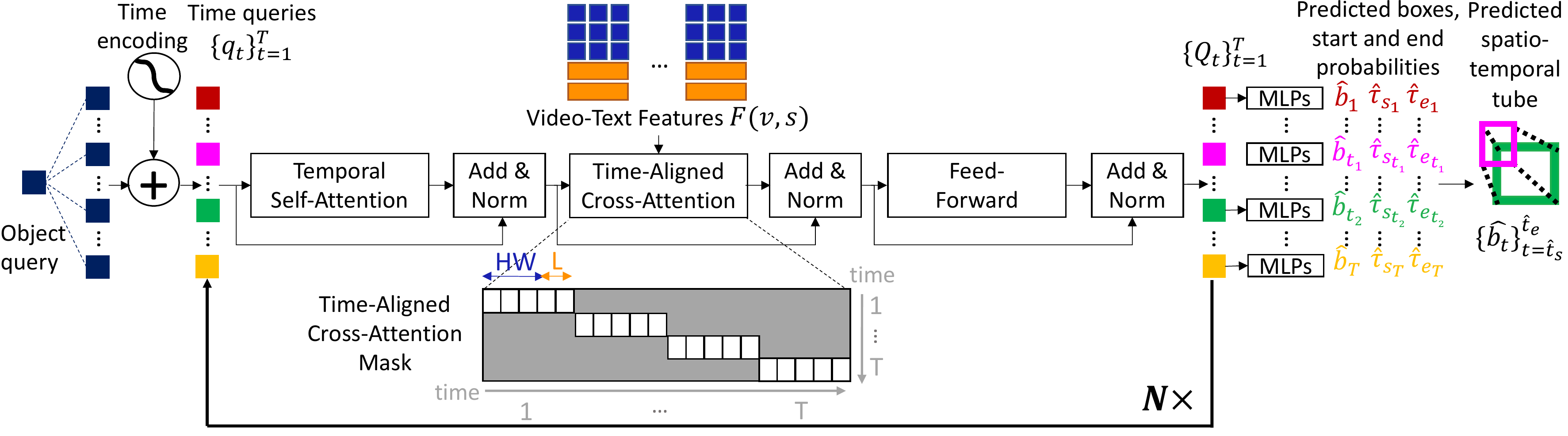} 
\vspace{-0.7cm}
\caption{\small \textbf{Space-Time Decoder}. 
The decoder is composed of $N$ repeated blocks. 
In each block, time queries $q_t$ successively attend to each other via {\em temporal self-attention} and to their respective time-aligned video-text features $F(v, s)$ via {\em time-aligned cross-attention}. 
The cross-attention mask (bottom) indicates the non-zero weights (white) between the input $HW+L$ video-language features for each of the $T$ input frames (x-axis) and $T$ time queries (y-axis). 
The cross-attention mask ensures that each time query $q_t$ only cross-attends to video-text features $F(v, s)$ at the corresponding frame $t$, which significantly increases efficiency of the decoder and enables decoding entire videos of $T$ frames. 
The temporal modelling over the entire length of the video is ensured by the temporal self-attention layers. 
}
\vspace{-0.5cm}
\label{fig:decoder}
\end{figure*}

\subsection{Space-Time Decoder}\label{sec:decoder}
Our decoder is illustrated in Figure~\ref{fig:decoder} and detailed next. 
Its objective is to model the temporal interactions within the entire video of $T$ frames and decode the multi-modal features from the encoder into a temporally coherent output tube with accurate start and end times. 
This is achieved by an efficient decoder architecture that alternates (i) {\em temporal self-attention} layers, which model {\em temporal} interactions across the entire video, with (ii) {\em time-aligned cross attention} layers, which efficiently incorporate the video-text features for individual frames obtained from the encoder.
In detail, the decoder operates on $T$ positional encodings $\{q_t\}_{t=1}^T$, one per frame, referred to as time queries. 
The initial encoding of each time query is obtained by summing a learnt object encoding common to all frames, and a frozen sinusoidal time encoding. 
The decoder also takes as input ${T \times (HW+L)}$ video-language embeddings $F(v, s)$ output from the video-text encoder.
The decoder is a succession of $N$ decoder blocks. 
Each block is composed of temporal self-attention, time-aligned cross-attention, and feed-forward layers, interleaved with normalization~\cite{ba2016layer}, as shown in Figure~\ref{fig:decoder}.
The decoder outputs refined time queries $\{Q_t\}_{t=1}^T$, which are contextualized across all frames in the video together with video-text features produced by the encoder. 
The refined time queries are then jointly used for outputting the spatio-temporal video tube that grounds the input sentence in the video. The individual layers are described in detail next.

\noindent \textbf{Temporal self-attention.}
The $T$ input time queries $q_t$ attend to each other using the temporal self-attention layer. 
This layer is in each of the $N$ blocks of the decoder and is responsible for modelling the long-range temporal interactions in the entire video.
This is possible because of the relatively low complexity of this layer, which does not depend on the spatial resolution of the input video. 

\noindent \textbf{Time-aligned cross-attention.} 
Allowing each time query to cross-attend to all ${T \times (HW+L)}$ video-text features can be highly computationally expensive due to the large number of video frames $T$ and a large spatial resolution $HW$ of the video features. 
Instead, in our cross-attention module, each time query $q_t$ only cross-attends to its temporally corresponding multi-modal features $F(v, s)[t]$ at frame $t$. 
Note that with our time-aligned cross-attention formulation, the time encoding and the temporal self-attention layers are all the more important, as they are responsible for the temporal modelling across the entire video.
Without them, our decoder would be decoding each frame independently. 
Their importance is ablated in Section~\ref{sec:ablations}.

\noindent \textbf{Prediction heads.}
The output of the decoder is a set of refined time queries $\{Q_t\}_{t=1}^T$. 
They are jointly used for visual grounding and temporal localization to simultaneously obtain predictions for {\em all frames of the video}.
In detail, normalized coordinates of all bounding boxes (2D center and size) $\hat{b} \in [0,1]^{T\times 4}$ are predicted with a 3-layer MLP.
Probabilities of the start and the end of the output video tube, $\hat{\tau}_s \in [0,1]^{T}$ and $\hat{\tau}_e \in [0,1]^{T}$, respectively, are predicted with 2-layer MLPs.
At inference time, the start and end times of the output tube, $\hat{t}_s$ and $\hat{t}_e$, are computed by choosing the maximum of the joint start and end probability distribution $(\hat{\tau}_s, \hat{\tau}_e) \in [0,1]^{T\times T}$ with invalid combinations where $\hat{t}_e \leq \hat{t}_s$ masked out.
The predicted spatio-temporal tube $\{\hat{b}_t\}_{t=\hat{t}_s}^{\hat{t}_e}$ is composed from bounding boxes $\hat{b}_t$ predicted within the chosen start and end times $\hat{t}_s$ and $\hat{t}_e$.

\subsection{Training loss}\label{sec:losses}
The input training data is in the form of a set of videos, where each video is annotated with a query sentence $s$ and the corresponding video tube $b$ composed of a set of bounding boxes and corresponding start and end times, $t_s$ and $t_e$.  
Inspired by~\cite{rodriguez2020proposal}, we construct a target start (respectively end) distribution $\tau_s \in [0,1]^{T}$ (respectively $\tau_e$) which follows a quantized Gaussian centered at  $t_s \in [0, T-1]$ (respectively $t_e$) with standard deviation 1.
We train our architecture with a linear combination of four losses \\[-17pt]
\begin{align}
\label{eq:objective}
\mathcal{L} = \lambda_{\mathcal{L}_1}\mathcal{L}_{\mathcal{L}_1}(\hat{b}, b) + \lambda_{gIoU}\mathcal{L}_{gIoU}(\hat{b}, b) \mspace{50mu}\notag\\
+ \lambda_{KL}\mathcal{L}_{KL}(\hat{\tau}_s, \hat{\tau}_e, \tau_s, \tau_e) 
+ \lambda_{att}\mathcal{L}_{att}(A)
\end{align}\\[-15pt]
\noindent where $b \in [0,1]^{4(t_e-t_s+1)}$ denotes the normalized ground truth box coordinates and $\hat{b}$ the predicted bounding boxes and $A \in [0,1]^{T\times T}$ denotes the temporal self-attention matrix.
Finally, different $\lambda_{\bullet}$ are scalar weights of the individual losses. 
$\mathcal{L}_{\mathcal{L}_1}$ is a $\mathcal{L}_1$ loss on bounding box coordinates. 
$\mathcal{L}_{gIoU}$ is a generalized ``intersection over union" (IoU) loss~\cite{Rezatofighi_2018_CVPR} on the bounding boxes. 
Both $\mathcal{L}_1$ and $\mathcal{L}_{gIoU}$ are used for spatial grounding. 
$\mathcal{L}_{KL}(\hat{\tau}_s, \hat{\tau}_e, \tau_s, \tau_e)$ is the Kullback-Leibler divergence loss measuring the distance between the predicted and the target start distribution as well as the distance between the predicted and the target end distribution~\cite{rodriguez2020proposal}.
$\mathcal{L}_{att}(A)$ is a guided attention loss~\cite{rodriguez2020proposal} that encourages weights corresponding to time queries outside of the temporal boundaries to be lower than the weights inside these boundaries. 
$\mathcal{L}_{KL}$ and $\mathcal{L}_{att}(A)$ are both used for temporal grounding. 
Losses are computed at each layer of the decoder following~\cite{carion2020end}.

\subsection{Weight initialization}\label{sec:weights}
We initialize our architecture with weights from MDETR~\cite{mdetr} pretrained on Flickr30k~\cite{plummer2015flickr30k}, MS COCO~\cite{chen2015microsoft} and Visual Genome~\cite{visualgenome}.
In detail, weights of our video-text encoder are initialized from the MDETR multi-modal encoder, except for the fast and aggregation modules.
We also use the weights from the MDETR single-image multi-object decoder to initialize our multi-frame single-object space-time decoder, except for the temporal localization head.
We show the benefit of this initialization notably by comparing it to an ImageNet initialization, \ie using a visual backbone pretrained on ImageNet with a randomly initialized transformer, in Section~\ref{sec:ablations}.
We also evaluate a MDETR-equivalent baseline in Section~\ref{sec:ablations}.

\section{Experiments}\label{sec:experiments}
This section demonstrates the effectiveness of our architecture and compares our method to the state of the art.
We first introduce the datasets, evaluation metrics and implementation details in Section \ref{sec:protocol}.
We then present ablation studies in Section \ref{sec:ablations}.
The comparison to the state of the art in spatio-temporal video grounding is given in Section \ref{sec:sota}.
Finally, we show qualitative results in Section \ref{sec:quali}.

\subsection{Experimental setup}\label{sec:protocol}

\paragraph{Datasets.}\label{sec:datasets}
We evaluate our approach on the VidSTG~\cite{vidstg} and HC-STVG~\cite{hcstvg} datasets. 
Both are annotated with spatio-temporal tubes corresponding to text queries.
\textbf{VidSTG}\label{sec:vidstg} consists of 99,943 sentence descriptions with 44,808 declarative sentences and 55,135 interrogative sentences describing 79 types of objects appearing in 10,303 different videos. 
The dataset is divided into training, validation and test subsets with 80,684, 8,956 and 10,303 distinct sentences respectively, and 5,436, 602 and 732 distinct videos respectively. 
\textbf{HC-STVG}\label{sec:hcstvg} consists of videos in multi-person scenes, each annotated with one sentence referring to a person. 
For ablation, we use the second improved version of the dataset~\textbf{HC-STVG2.0} which is divided into training and validation subsets with 10,131 and 2,000 video-sentence pairs, respectively. 
The test set is not publicly available at the time of writing. 
To compare with prior work, we use the first version of the dataset~\textbf{HC-STVG1} which is divided into training and test subsets with 4,500 and 1,160 video-sentence pairs, respectively. 

\vspace{-.4cm}
\paragraph{Evaluation metrics.}\label{sec:metrics} We follow \cite{vidstg} and define $vIoU$ as $vIoU=\frac{1}{|S_u|}\sum_{t \in S_i}IoU(\hat{b}_t, b_t)$ where $S_u$ (respectively $S_i$) is the set of frames in the union (respectively intersection) between the ground truth (GT) and the predicted timestamps.
$\hat{b}_t$ (respectively $b_t$) are the predicted (respectively GT) boxes at time t. 
To evaluate spatio-temporal video grounding, we use $m\_vIoU$, which is the average of $vIoU$. We also use $vIoU@R$, the proportion of samples for which $vIoU>R$. 
To isolate the evaluation of temporal localization, we use $m\_tIoU$ which is the average of temporal IoU between the GT start and end and the predicted start and end.
Likewise, to evaluate spatial grounding only, we use $\siou{}$, which is computed by using the GT start and end times.
For ablations we report results averaged over all samples. 
More detailed ablation results presented separately for declarative and interrogative sentences in VidSTG are reported in Appendix Section~\ref{sec:addablation}. 
We also report peak GPU memory usage during training (Mem.) to measure the memory footprint of alternative models.

\vspace{-.4cm}
\paragraph{Implementation details.}\label{sec:details}
The visual backbone is ResNet-101~\cite{he16resnet}, the text encoder is RoBERTa~\cite{liu2019roberta} and the fast module $f$ is a linear layer. 
Following \cite{vidstg}, we sample 5 frames per second for videos, and for videos with more than 200 sampled frames we uniformly sample 200 frames.
We use hyper-parameters $T=200$, $N=6$, $d=256$, $\lambda_{\mathcal{L}_1} = 5$, $\lambda_{giou} = 2$, $\lambda_{KL} = 10$ and $\lambda_{att} = 1$. 
We train our networks for 10, 20 and 40 epochs on VidSTG, HC-STVG2.0 and HC-STVG1, respectively. 
The final model is selected based on the best spatio-temporal video grounding performance on the validation set.
For the largest dataset VidSTG, the optimization takes 2 days on 16 Tesla V100 GPUs.
Further details are included in Appendix Section~\ref{sec:adddetails}. 

\subsection{Ablation studies}\label{sec:ablations}

\begin{table}[t]
\centering
\vspace{-0pt}
\begin{center}
\setlength\tabcolsep{4pt}
\resizebox{1.\linewidth}{!}{
\begin{tabular}{ccc|ccccc}
& \makecell{\small{Time} \\ \small{Encoding}} & \makecell{\small{Self} \\ \small{Attention}} & \small{m\_tIoU} & \small{m\_vIoU} & \makecell{\small{vIoU} \\ \small{@0.3}} & \makecell{\small{vIoU} \\ \small{@0.5}}
& \small{\siou{}} \\ 
\hline
1. & \xmark & - & 23.9 & 12.2
& 15.3 & 6.1 & 47.0 \\ 
2. & \xmark & Temporal & 25.2 & 13.0
& 16.9 & 6.5 & 47.3 \\ 
3. & \cmark & - & 41.7 & 21.3
& 28.7 & 17.4 & 46.5 \\ 
4. & \cmark & Temporal & \textbf{45.9} & \textbf{24.3} 
& \textbf{33.2} & \textbf{22.0} & \textbf{47.7} \\ 
\end{tabular}}
\vspace{-0.4cm}
\caption{\small Effect of the time encoding and the temporal self-attention in our space-time decoder on the VidSTG validation set.} 
\vspace{-0.7cm}
\label{table:decoder}
\end{center}
\end{table}

\begin{table}[t]
\vspace{-0pt}
\begin{center}
\setlength\tabcolsep{3pt}
\resizebox{1.\linewidth}{!}{
\begin{tabular}{ccc|ccccc}
& \makecell{\small{Pre-} \\ \small{Training}} & \makecell{\small{Decoder Self-} \\ \small{Attention Transfer}} & \small{m\_tIoU} & \small{m\_vIoU} & \makecell{\small{vIoU} \\ \small{@0.3}} & \makecell{\small{vIoU} \\ \small{@0.5}}
& \small{\siou{}} \\ 
\hline
1. & \xmark & \xmark & 42.8 & 18.8
& 25.1 & 15.6 & 38.5 \\ 
2. & \cmark & \xmark & 43.8 & 22.4
& 29.9 & 19.1 & 46.5 \\ 
3. & \cmark & Temporal & \textbf{45.9} & \textbf{24.3} 
& \textbf{33.2} & \textbf{22.0} & \textbf{47.7} \\ 
\end{tabular}}
\vspace{-0.4cm}
\caption{\small Effect of the weight initialization for our model on the VidSTG validation set.}
\vspace{-1cm}
\label{table:init}
\end{center}
\end{table}

\begin{table*}
\begin{center}
\begin{subtable}[t]{0.49\textwidth}
\caption{\small VidSTG}
\vspace{-0.6cm}
\begin{center}
\setlength\tabcolsep{2pt}
\resizebox{1.\linewidth}{!}{
\begin{tabular}{llll|ccccc|c}
& Fast & Res. & \makecell{ \small{Temp.} \\ \small{Stride}} & \small{m\_tIoU} & \small{m\_vIoU} & \small{vIoU@0.3} & \small{vIoU@0.5} & \small{\siou{}} 
& \makecell{\small{Mem.} \\ \small{(GB)}} \\ 
\hline
1. & --- & 224 & 1 & 46.5 & 25.2 & 34.1 & 23.0 & 49.1 & 23.9 \\
2. & \cmark & 224 & 2 & 46.0 & 25.0 & 34.3 & 22.9 & 49.0 & 16.2 \\
3. & \cmark & 224 & 5 & 45.9 & 24.3 & 33.2 & 22.0 & 47.7 & 11.8 \\
4. & \cmark & 288 & 2 & 46.4 & 25.9 & 35.0 & 23.9 & 50.5 & 23.7 \\
5. & \cmark & 320 & 3 & 46.4 & 25.9 & 35.7 & 23.7 & \textbf{50.7} & 23.6 \\
6. & \cmark & 352 & 4 & \textbf{46.9} & \textbf{26.2} & \textbf{36.1} &\textbf{24.1} & \textbf{50.7} & 24.4 \\
7. & \xmark & 352 & 4 & 46.6 & 24.8 & 34.0 & 21.6 & 48.3 & 18.1 \\
8. & \cmark & 384 & 5 & 46.8 & 26.0 & 35.5 & 24.0 & 50.4 & 26.1 \
\end{tabular}}
\label{table:resolution}
\end{center}
\end{subtable}
\hfill
\begin{subtable}[t]{0.49\textwidth}
\caption{\small HC-STVG2.0}
\vspace{-0.6cm}
\begin{center}
\setlength\tabcolsep{2pt}
\resizebox{1.\linewidth}{!}{
\begin{tabular}{llll|ccccc|c}
& Fast & Res. & \makecell{ \small{Temp.} \\ \small{Stride}} & \small{m\_tIoU} & \small{m\_vIoU} & \small{vIoU@0.3} & \small{vIoU@0.5} &
\small{\siou{}} & \makecell{\small{Mem.} \\ \small{(GB)}} \\ 
\hline
1. & --- & 224 & 1 & 52.8 & 35.0 & 55.3 & 28.3 & 63.9 & 14.3 \\
2. & \cmark & 224 & 2 & 53.7 & 35.8 & 56.7 & 29.6 & 64.3 & 10.2 \\
3. & \cmark & 224 & 5 & 53.2 & 35.0 & 54.5 & 29.0 & 63.2 & 8.0 \\
4. & \cmark & 288 & 2 & \textbf{53.9} & \textbf{36.4} & 58.1 & \textbf{30.7} & \textbf{65.4} & 13.9 \\
5. & \cmark & 320 & 3 & 53.6 & 36.2 & 57.5 & 30.4 & 65.2 & 13.8 \\
6. & \cmark & 352 & 4 & \textbf{53.9} & \textbf{36.4} & \textbf{58.8} & 30.6 & 64.9 & 14.3 \\
7. & \xmark & 352 & 4 & 53.1 & 34.7 & 55.9 & 27.4 & 63.0 & 11.3 \\
8. & \cmark & 384 & 5 & 53.6 & 36.3 & 57.5 & 30.4 & 65.3 & 15.2 \\
\end{tabular}}
\label{table:resolution2}
\end{center}
\end{subtable}
\vspace{-0.4cm}
\caption{\small Comparison of performance-memory trade-off with various temporal strides $k$, spatial resolutions (Res.), with or without the fast branch in our video-text encoder, on the VidSTG validation set (left, Table~\ref{table:resolution}) and the HC-STVG2.0 validation set (right, Table~\ref{table:resolution2}).}
\label{table:resolutions}
\vspace{-0.7cm}
\end{center}
\end{table*}

\begin{table*}[t]
\begin{center}
\setlength\tabcolsep{1pt}
\resizebox{1\linewidth}{!}{
\begin{tabular}{llc|cccc|cccc|ccc}
& \multirow{3}{*}{Method} & 
\multirow{3}{*}{\makecell{\small{Pretraining} \\ \small{Data}}} 
& \multicolumn{8}{c|}{VidSTG}
& \multicolumn{3}{c}{HC-STVG1} \\ 
& & & \multicolumn{4}{c}{Declarative Sentences} & \multicolumn{4}{c|}{Interrogative Sentences} \\ 
& & & m\_tIoU & m\_vIoU & \small{vIoU@0.3} & \small{vIoU@0.5}
& m\_tIoU & m\_vIoU & \small{vIoU@0.3} & \small{vIoU@0.5} 
& m\_vIoU & vIoU@0.3 & vIoU@0.5 \\
\hline
1. & STGRN~\cite{vidstg} & Visual Genome
& \textbf{48.5} & 19.8 & 25.8 & 14.6 
& \textbf{47.0} & 18.3 & 21.1 & 12.8 
& --- & --- & --- \\ 
2. & STGVT~\cite{hcstvg} & \makecell{\small{Visual Genome +} \\[-2pt] \small{Conceptual Captions}}
& --- & 21.6 & 29.8 & 18.9
& --- & --- & --- & --- 
& 18.2 & 26.8 & 9.5 \\ 
3. & STVGBert~\cite{stvgbert} & \makecell{\small{ImageNet + Visual Genome +} \\[-2pt] \small{Conceptual Captions}}
& --- & 24.0 & 30.9 & 18.4 
& --- & 22.5 & 26.0 & 16.0 
& 20.4 & 29.4 & 11.3 \\ 
\hline
4. & TubeDETR (Ours) & ImageNet & 43.1 & 22.0 & 29.7 & 18.1
& 42.3 & 19.6 & 26.1 & 14.9 
& 21.2 & 31.6 & 12.2 \\
5. & TubeDETR (Ours) & \makecell{\small{ImageNet + Visual Genome +} \\[-2pt] \small{Flickr + COCO}}
& 48.1 & \textbf{30.4} & \textbf{42.5} & \textbf{28.2}
& 46.9 & \textbf{25.7} & \textbf{35.7} & \textbf{23.2}
& \textbf{32.4} & \textbf{49.8} & \textbf{23.5} \\
\end{tabular}
}
\end{center}
\vspace{-0.7cm}
\caption{\small Comparison to the state of the art on the VidSTG test set and the HC-STVG1 test set.}  
\vspace{-0.7cm}
\label{table:sota}
\end{table*}

In this section, we ablate the hyper-parameters of our model and evaluate alternative design choices of the encoder and decoder. 
Unless stated otherwise, we use spatial frame resolution of 224 pixels and temporal stride $k=5$.

\noindent \textbf{Space-time decoder.}\label{sec:decoderres}
We first ablate the design choices of the proposed space-time decoder. 
We compare our full decoder model with variants without time encoding, without temporal self-attention and without both.
The variant without both corresponds to a space-only decoder, similar to MDETR~\cite{mdetr} applied independently to every frame.
Table~\ref{table:decoder} shows that there is a substantial improvement over the space-only decoder when using both time encoding and temporal self-attention (+17.9\% on $vIoU@0.3$ between rows 1 and 4).
The gain comes mostly from the temporal localization (+22.0\% on $m\_tIoU$), while the spatial grounding moderately increases (+0.7\% in $\siou{}$). 
Furthermore, we can observe that the time encoding brings most of the gain (+13.4\% on $vIoU@0.3$ between rows 1 and 3).
Finally, the temporal self-attention results in an additional improvement (+4.5\% on $vIoU@0.3$ between rows 3 and 4) over using time encoding only. 

\noindent \textbf{Initialization.}\label{sec:init}
We now ablate the importance of initializing our model with pretrained MDETR~\cite{mdetr} weights.
In Table~\ref{table:init}, we compare this initialization to ImageNet initialization, and a variant that does not transfer the spatial self-attention weights from MDETR decoder to the temporal self-attention in our space-time decoder.
At pretraining time, this self-attention was used to model spatial relationships between different objects in the same image, while the temporal self-attention in our decoder models temporal relationships between the same object in different frames of a video.
We find that pretraining is highly beneficial (+8.1\% on $vIoU@0.3$ between rows 1 and 3), especially for the spatial grounding performance (+9.2\% on $\siou{}$).
Additionally, we observe the benefit of using the spatial self-attention weights from the MDETR decoder to initialize the temporal self-attention in our decoder (+3.3\% on $vIoU@0.3$ between rows 2 and 3).

\noindent \textbf{Impact of spatial resolution and temporal stride $k$.}\label{sec:resolution}
In this section, we analyze the impact of the frame resolution and the temporal stride $k$.
In Table~\ref{table:resolutions}, we show that increasing the resolution is an important factor of performance for spatio-temporal video grounding, on both the VidSTG and HC-STVG2.0 datasets (see rows 2 and 4).
However, it also results in significantly higher memory usage (16.2GB vs 23.7GB).
As a consequence, the variant using temporal stride $k=1$ is challenging to train on VidSTG with a resolution higher than 224 on a Tesla V100 32GB GPU.
At a fixed 224 resolution, increasing the temporal stride $k$ to 2 or 5 reduces the peak memory usage by 7.7GB or 12.1GB, respectively (see row 1 vs 2 or 3, respectively).
Our proposed video-text encoder enables us to train on higher resolutions at a given memory usage.
This leads to a better performance-memory trade-off (rows 4, 5, 6, 8) than the baseline variant with temporal stride $k=1$ (row 1).
In particular, the best spatio-temporal video grounding results ($m\_vIoU$ and $vIoU@R$) over the two datasets are obtained with temporal stride $k=4$ and resolution 352 (row 6).

We note that as the resolution increases, performance gains obtained by its further increase are expected to be lower as they are limited by the original video resolution.
For instance, the average video pixel height in VidSTG and HCSTVG2.0 is 440 and 490 pixels, respectively.

\noindent \textbf{Impact of the fast branch.} 
Finally, we validate the importance of our fast branch by comparing, for the best variant, temporal stride $k=4$ and resolution 352, our slow-fast video-text encoder to a slow-only variant that corresponds to $f=0$ and $g=0$.
In this case the video-text features are the slow video-text features.
By comparing rows 6 and 7 in Table~\ref{table:resolutions}, our fast branch significantly improves the spatio-temporal video grounding performance (+2.1\% $vIoU@0.3$ on VidSTG and +2.9\% $vIoU@0.3$ on HC-STVG2.0) with low computational memory overhead.
This shows that the fast branch recovers useful spatio-temporal details lost by the temporal sampling operation in the slow branch.
We further ablate the design of the fast and aggregation modules $f$ and $g$ in Appendix Section~\ref{sec:addexp}. 

\begin{figure*}[t]
\centering
\includegraphics[width=1.\linewidth]{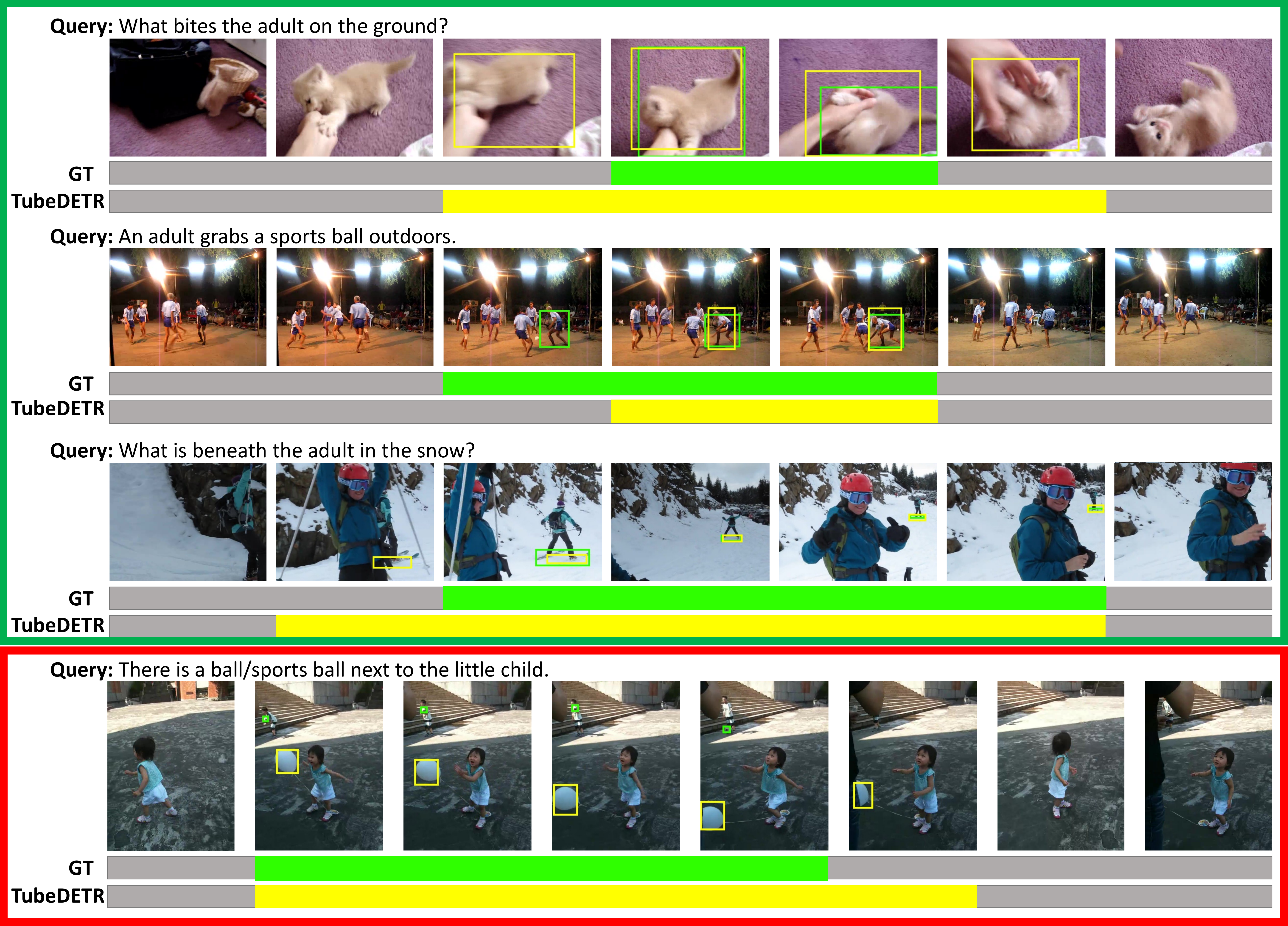}
\caption{\small Qualitative examples of spatio-temporal tubes predicted by our model (light yellow), compared with ground truth (light green), on the VidSTG test set. 
The first three examples illustrate successful predictions of our method.
In the last example the method confuses the small sports ball in the background with a balloon.}
\label{fig:qualitative}
\end{figure*}

\subsection{Comparison to the state of the art}\label{sec:sota}
In this section, we compare our approach to state-of-the-art methods in spatio-temporal video grounding. 
We report results for the model achieving the best validation results in the previous ablation studies, \ie, our space-time decoder with time encoding and temporal self-attention, temporal stride $k=4$ and resolution 352. 
The focus of our work is on the spatio-temporal video grounding metrics ($m\_vIoU$ and $vIoU@R$).
As shown in Table~\ref{table:sota}, only using  ImageNet to initialize the visual backbone (row 4), our TubeDETR performs competitively despite using less annotations.
Furthermore, if we use MDETR initialization (row 5), our TubeDETR outperforms by a large margin all previous methods (rows 1, 2 and 3) on both datasets.
STGRN~\cite{vidstg} achieves similar $m\_tIoU$ (measuring only temporal localization), but it defines a handcrafted set of possible window widths to tackle temporal localization, while we consider all possible windows, \ie any starting frame $i$ and ending frame $j$ with $i<j$.
These results demonstrate the excellent performance of our architecture for spatio-temporal video grounding.

\subsection{Qualitative examples}\label{sec:quali}
We show qualitative examples of our predictions on the VidSTG test set in Figure~\ref{fig:qualitative}. 
These examples show that our model is able to predict meaningful and accurate spatio-temporal tubes associated with the input text queries.
In particular, in the first example, our model correctly detects the temporal moment corresponding to the cat biting the adult.
In the second example, our model localizes  the spatio-temporal tube corresponding to a man quickly grabbing a very small sports ball and in the third example it is able to localize the skis under the adult while skiing.
However, as shown in the last example, it may fail to understand fine details in the query and the video.
Note that the balloon and the ball are visually and semantically similar. 
A careful analysis is required to understand the difference.
Furthermore, we provide visualizations of the different attention mechanisms of TubeDETR in Appendix Section~\ref{sec:viz}. 

\section{Conclusion}\label{sec:conclusion}
We have proposed TubeDETR, a novel transformer-based architecture for spatio-temporal video grounding.
TubeDETR tackles this task with a space-time transformer decoder combined with a video-text encoder that efficiently encodes spatial and multi-modal interactions.
We have demonstrated the effectiveness of our space-time decoder, and the benefits of our video-text encoder in terms of performance-memory trade-off.
Finally, our approach outperforms state-of-the-art methods on two benchmarks, VidSTG and HC-STVG.
Future work could extend our space-time decoder to detect multiple objects per frame or multiple events per video.
Investigating more efficient alternatives to self-attention, such as the ones studied for natural language~\cite{kitaev2020reformer, wu2020lite, wang2020linformer, beltagy2020longformer, tay2020long, choromanski2020rethinking, zaheer2020big}, is another promising direction for future research.

\mbox{}\vspace{-0.3cm}\\
\noindent 
{
\footnotesize{
{\textbf{Acknowledgements.} This work was granted access to the HPC resources of IDRIS under the allocation 2021-AD011011670R1 made by GENCI. The work was funded by a Google gift,  the French government under management of Agence Nationale de la Recherche as part of the "Investissements d'avenir" program, reference ANR-19-P3IA-0001 (PRAIRIE 3IA Institute), the Louis Vuitton ENS Chair on Artificial Intelligence, the European Regional Development Fund under project IMPACT (reg.\ no.\ CZ.02.1.01/0.0/0.0/15 003/0000468).
We thank S. Chen and J. Chen for helpful discussions and O. Bounou and P.-L. Guhur for proofreading.}
}}

{\small
\bibliographystyle{ieee_fullname}
\bibliography{egbib}
}

\clearpage \newpage
\appendix

\section*{Appendix}
In this Appendix, we present additional visualizations of the different attention mechanisms in our space-time decoder in Section~\ref{sec:viz}. 
Section~\ref{sec:adddetails} provides additional implementation details. 
We then give detailed results for ablations in Section~\ref{sec:ablations} on the VidSTG dataset~\cite{vidstg} split by sentence type in Section~\ref{sec:addablation}.
Next we present an ablation of our fast and aggregation modules in Section~\ref{sec:addexp}.
Finally we discuss broader impact in Section~\ref{sec:impact}.

\section{Visualization of space, time and language attention patterns in the decoder}\label{sec:viz}
\begin{figure}[t]
\centering
\includegraphics[width=1.\linewidth]{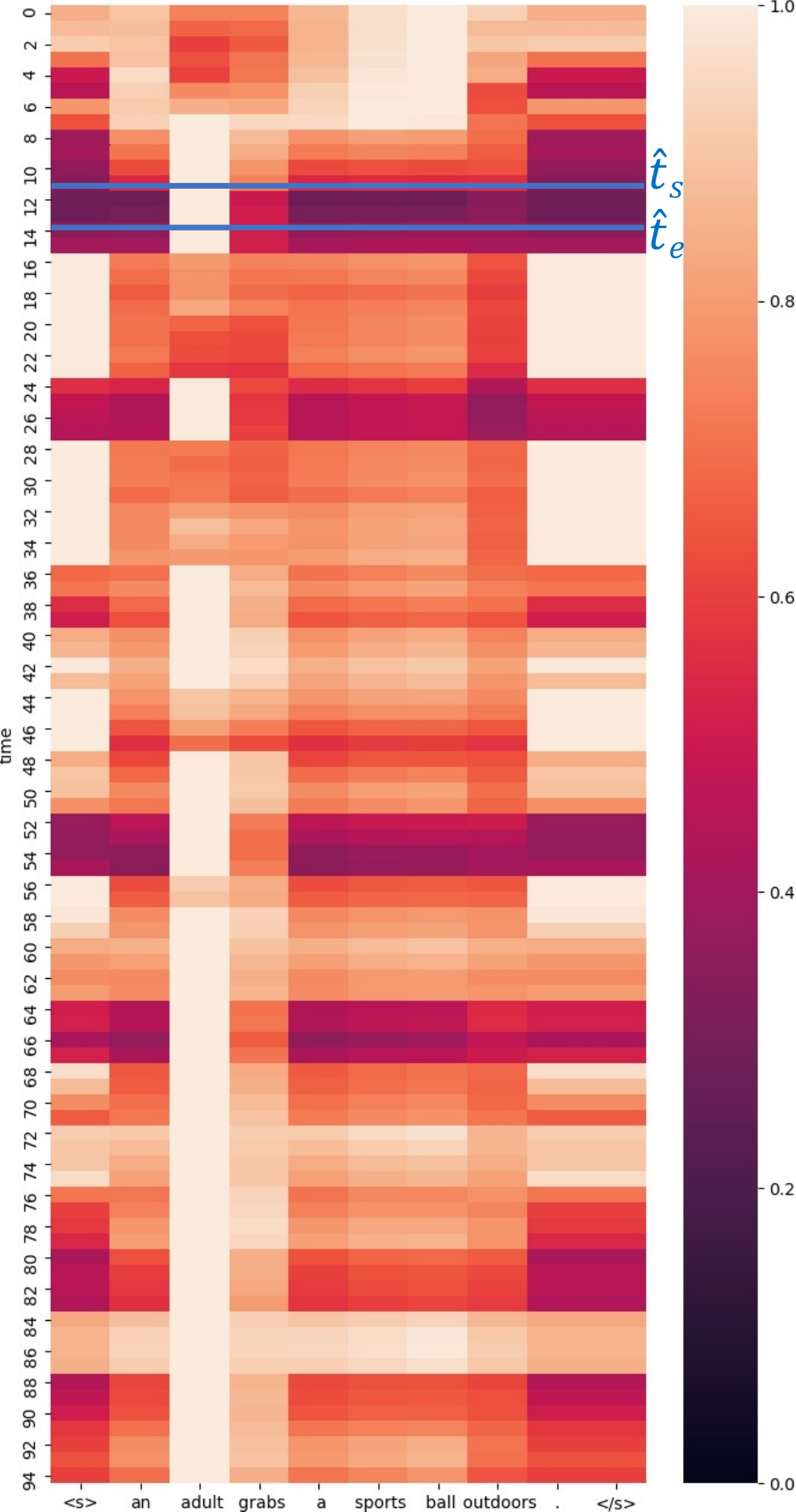}
\vspace{+0.1cm}
\caption{\small \textbf{Time-aligned cross-attention visualization (textual modality).} 
Visualization of the attention weights between the time query (y-axis) and its time-aligned visually-contextualized text features (x-axis) at different times in our space-time decoder. 
These attention weights are averaged across all 8 heads and all 6 layers, and renormalized by the maximum weight at each timestep (\ie each row) for the purpose of visualization. Lighter colors correspond to higher attention weights (see the colorbar on the right).
}
\vspace{+0.1cm}
\label{fig:textattn}
\end{figure}

\begin{figure*}[t]
\centering
\includegraphics[width=1\linewidth]{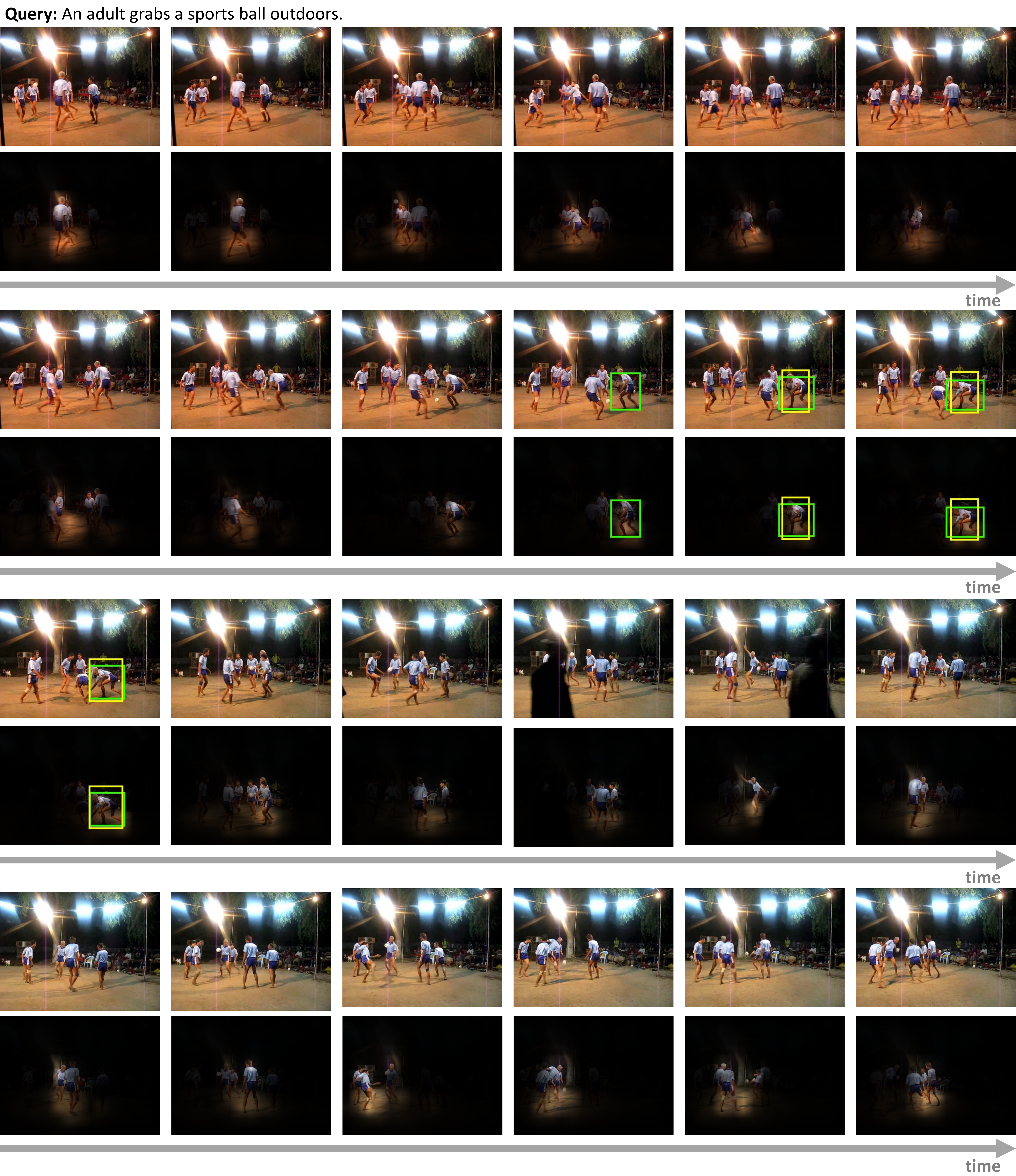} 
\caption{\small \textbf{Time-aligned cross-attention visualization (visual modality).} 
Top rows: Input frames with the predicted (yellow) and ground truth (green) spatio-temporal tubes overlaid.
Bottom rows: Visualization of the attention weights between the time query and its time-aligned text-contextualized visual features at different times in our space-time decoder. 
These attention weights are averaged across all 8 heads and all 6 layers, and renormalized by the maximum weight at each timestep for the purpose of visualization.
Attention at each timestep is particularly focused on humans that are receiving the sports ball and gesturing.
}
\label{fig:spaceattn}
\vspace{-1cm}
\end{figure*}

\begin{figure*}[!htbp]
\centering
\includegraphics[width=1\linewidth]{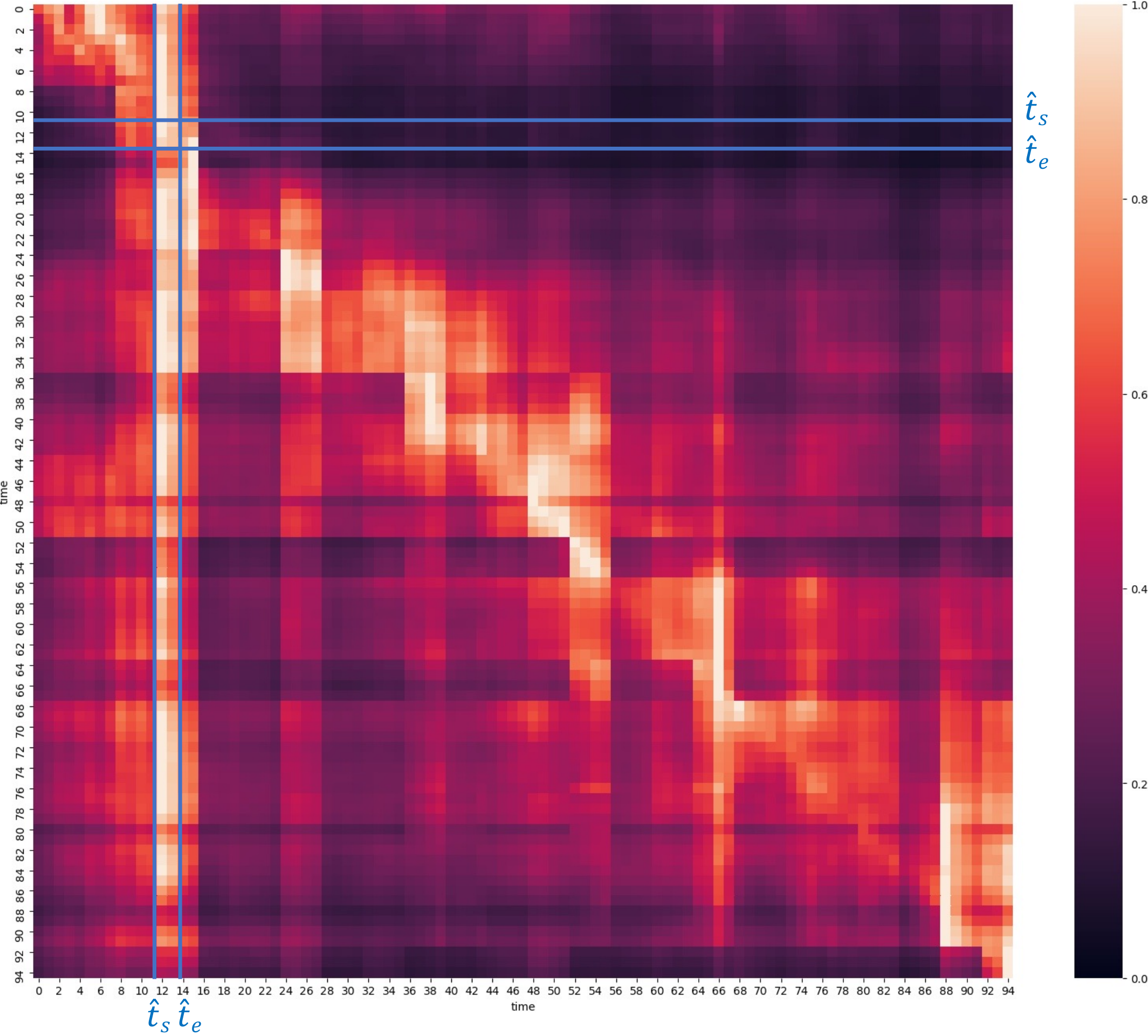}
\caption{\small \textbf{Temporal self-attention visualization.}
Visualization of the attention weights between the different time queries in our space-time decoder. 
The column $t$ corresponds to the weights of the different time queries for the time query at time $t$.
These attention weights are averaged across all 8 heads and all 6 layers, and renormalized by the maximum weight at each timestep (\ie each column) for the purpose of visualization.
$\hat{t}_s$ and $\hat{t}_e$ denote the predicted start and end times of the output tube.
Lighter colors correspond to higher attention weights (see the colorbar on the right).
}
\label{fig:timeattn}
\end{figure*}

This section illustrates attention mechanisms of our space-time decoder over space, language and time for the spatio-temporal video grounding example presented in Figure~\ref{fig:spaceattn}. 
For this example the time-aligned cross-attention for the visual modality is also shown in Figure~\ref{fig:spaceattn}.
We note that spatially, attention at each timestep is particularly focused on humans that are receiving the sports ball and gesturing.
Additionally, the time-aligned cross-attention for the textual modality is illustrated in Figure~\ref{fig:textattn}. 
We observe that the words \textit{adult} and \textit{grabs} are the most attended overall, and that attention weights on the different words (\eg \textit{sports} and \textit{ball}) vary over time. $\hat{t}_s$ and $\hat{t}_e$ in Figure~\ref{fig:textattn} denote the predicted start and end times of the output tube.
Next, the temporal self-attention is illustrated in Figure~\ref{fig:timeattn}. 
We notice long-range temporal interactions: a certain number of time queries attend to various temporally distant time queries, \eg time queries located around the start of the video between the eighth and sixteenth frames.

\section{Additional implementation details}\label{sec:adddetails}

In our transformer, the number of heads is 8 and the hidden dimension of the feed-forward layers is 2048.
We set the initial learning rates to $1e^{-5}$ for the visual backbone, and $5e^{-5}$ for the rest of the network.
The learning rate follows a linear schedule with warm-up for the text encoder and the learning rate is constant for the rest of the network.
We use the AdamW optimizer~\cite{loshchilov2017decoupled} and weight-decay $1e^{-4}$.
Video data augmentation includes spatial random resizing, spatial random cropping preserving box annotations, and temporal random cropping preserving the annotated time interval.
Dropout~\cite{srivastava2014dropout} with probability 0.1 is applied in our transformer layers, and dropout with probability 0.5 is applied in the temporal localization head.
We use exponential moving average with a decay rate of 0.9998, and an effective batch size of 16 videos.
For temporal stride $k=1$ the fast and aggregation modules in the encoder are not active, as their goal is to recover local spatial and temporal information when $k>1$.

\begin{table*}[t]
\centering
\vspace{-0pt}
\begin{center}
\setlength\tabcolsep{6pt}
\resizebox{.9\linewidth}{!}{
\begin{tabular}{ccc|ccccc|ccccc}
& \multirow{2}{*}{\makecell{\small{Time} \\ \small{Encoding}}} & \multirow{2}{*}{\makecell{\small{Self} \\ \small{Attention}}} & \multicolumn{5}{c}{Declarative Sentences} & \multicolumn{5}{c}{Interrogative Sentences} \\
& & & \small{m\_tIoU} & \small{m\_vIoU} & \makecell{\small{vIoU} \\ \small{@0.3}} & \makecell{\small{vIoU} \\ \small{@0.5}}
& \small{\siou{}} 
& \small{m\_tIoU} & \small{m\_vIoU} & \makecell{\small{vIoU} \\ \small{@0.3}} & \makecell{\small{vIoU} \\ \small{@0.5}}
& \small{\siou{}} \\ 
\hline
1. & \xmark & - 
& 24.4 & 13.6 & 17.8 & 7.3 & 51.9
& 23.5 & 11.1 & 13.3 & 5.2 & 43.1 \\ 
2. & \xmark & Temporal 
& 25.3 & 14.1 & 18.6 & 7.3 & 52.3
& 25.0 & 12.1 & 15.4 & 5.9 & 43.3 \\ 
3. & \cmark & - 
& 42.1 & 23.2 & 31.8 & 19.5 & 51.3
& 41.5 & 19.7 & 26.2 & 15.8 & 42.5 \\ 
4. & \cmark & Temporal 
& \textbf{46.4} & \textbf{26.6} & \textbf{36.1} & \textbf{24.7} & \textbf{52.8} 
& \textbf{45.6} & \textbf{22.5} & \textbf{30.8} & \textbf{19.8} & \textbf{43.6} \\ 
\end{tabular}}
\vspace{-0.3cm}
\caption{\small Effect of the time encoding and the temporal self-attention in our space-time decoder on the VidSTG validation set.}
\vspace{-0.5cm}
\label{table:adddecoder}
\end{center}
\end{table*}

\begin{table*}[t]
\vspace{-0pt}
\begin{center}
\setlength\tabcolsep{5pt}
\resizebox{.9\linewidth}{!}{
\begin{tabular}{ccc|ccccc|ccccc}
& \multirow{2}{*}{\makecell{\small{Pre-} \\ \small{Training}}} & \multirow{2}{*}{\makecell{\small{Decoder Self-} \\ \small{Attention Transfer}}} & \multicolumn{5}{c}{Declarative Sentences} & \multicolumn{5}{c}{Interrogative Sentences} \\
& & & \small{m\_tIoU} & \small{m\_vIoU} & \makecell{\small{vIoU} \\ \small{@0.3}} & \makecell{\small{vIoU} \\ \small{@0.5}}
& \small{\siou{}}
& \small{m\_tIoU} & \small{m\_vIoU} & \makecell{\small{vIoU} \\ \small{@0.3}} & \makecell{\small{vIoU} \\ \small{@0.5}}
& \small{\siou{}} \\ 
\hline
1. & \xmark & \xmark 
& 42.9 & 19.8 & 26.7 & 16.8 & 41.1
& 42.8 & 18.0 & 23.9 & 14.6 & 36.5 \\ 
2. & \cmark & \xmark 
& 44.0 & 24.5 & 32.9 & 21.5 & 51.5
& 43.6 & 20.8 & 27.5 & 17.2 & 42.6 \\ 
3. & \cmark & \cmark 
& \textbf{46.4} & \textbf{26.6} & \textbf{36.1} & \textbf{24.7} & \textbf{52.8} 
& \textbf{45.6} & \textbf{22.5} & \textbf{30.8} & \textbf{19.8} & \textbf{43.6} \\ 
\end{tabular}}
\vspace{-0.3cm}
\caption{\small Effect of the weight initialization for our model on the VidSTG validation set.}
\vspace{-0.5cm}
\label{table:addinit}
\end{center}
\end{table*}

\begin{table*}[!htbp]
\begin{center}
\setlength\tabcolsep{2pt}
\resizebox{.9\linewidth}{!}{
\begin{tabular}{llll|ccccc|ccccc|c}
& \multirow{2}{*}{Fast} & \multirow{2}{*}{Res.} & \multirow{2}{*}{\makecell{ \small{Temp.} \\ \small{Stride}}} & \multicolumn{5}{c}{Declarative Sentences} & \multicolumn{5}{c}{Interrogative Sentences} 
& \multirow{2}{*}{\makecell{\small{Mem.} \\ \small{(GB)}}} \\
& & & & \small{m\_tIoU} & \small{m\_vIoU} & \small{vIoU@0.3} & \small{vIoU@0.5}
& \small{\siou{}} & \small{m\_tIoU} & \small{m\_vIoU} & \small{vIoU@0.3} & \small{vIoU@0.5} &
\small{\siou{}} & \\
\hline
1. & --- & 224 & 1 
& 46.9 & 27.6 & 37.7 & 25.7 & 54.2 
& 46.1 & 23.3 & 31.3 & 20.8 & 44.9 & 23.9 \\
2. & \cmark & 224 & 2 
& 46.6 & 27.4 & 38.0 & 25.7 & 54.3 &
45.5 & 23.0 & 31.3 & 20.7 & 44.7 & 16.2 \\
3. & \cmark & 224 & 5 
& 46.4 & 26.6 & 36.1 & 24.7 & 52.8 & 
45.6 & 22.5 & 30.8 & 19.8 & 43.6 & 11.8 \\
4. & \cmark & 288 & 2 
& 47.0 & 28.2 & 38.3 & 26.3 & 55.7 & 
46.0 & 24.1 & 32.4 & \textbf{22.0} & 46.3 & 23.7 \\
5. & \cmark & 320 & 3 
& 46.9 & 28.3 & 39.2 & 26.4 & 56.0 & 
45.9 & 24.0 & 32.8 & 21.5 & \textbf{46.4} & 23.6 \\
6. & \cmark & 352 & 4 
& 47.2 & \textbf{28.7} & \textbf{39.6} & \textbf{27.1} & \textbf{56.4}
& \textbf{46.6} & \textbf{24.2} & \textbf{33.2} & 21.7 & 46.2 & 24.4 \\
7. & \xmark & 352 & 4 
& 47.1 & 27.1 & 37.4 & 24.1 & 53.7
& 46.2 & 22.9 & 31.3 & 19.6 & 44.0 & 18.1 \\
8. & \cmark & 384 & 5 
& \textbf{47.4} & 28.4 & 38.9 & 27.0 & 55.3 & 
46.4 & 24.0 & 32.8 & 21.7 & 45.6 & 26.1 \\
\end{tabular}}
\vspace{-0.3cm}
\caption{\small Comparison of performance-memory trade-off with various temporal strides $k$, frame spatial resolutions (Res.), with or without the fast branch in our video-text encoder, on the VidSTG validation set.}
\label{table:addresolution}
\end{center}
\vspace{-0.7cm}
\end{table*}

\section{Detailed ablation results}~\label{sec:addablation}
In this section, we provide detailed results split by sentence type (declarative, interrogative) on the VidSTG dataset for the ablation studies presented in Section~\ref{sec:ablations}.

\noindent \textbf{Space-time decoder.}\label{sec:adddecoderres}
We first provide detailed results for the ablation on our space-time decoder. 
The analysis is similar for both declarative and interrogative sentences.
In detail, Table~\ref{table:adddecoder} shows that there is a substantial improvement over the space-only decoder when using both time encoding and temporal self-attention (+18.3\% on $vIoU@0.3$ for declarative sentences and +17.5\% on $vIoU@0.3$ for interrogative sentences between rows 1~and 4). 
The gain comes mostly from the temporal localization (+22.0\% on $m\_tIoU$ for declarative sentences and +22.1\% on $m\_tIoU$ for interrogative sentences), while the spatial grounding moderately increases (+0.9\% in $\siou{}$ for declarative sentences and +0.5\% in $\siou{}$ for interrogative sentences). 
Furthermore, we can observe that the time encoding brings most of the gain (+14.0\% on $vIoU@0.3$ for declarative sentences and +12.9\% on $vIoU@0.3$ for interrogatives sentences between rows 1 and 3).
Finally, the temporal self-attention results in an additional improvement (+4.3\% on $vIoU@0.3$ for declarative sentences and +4.6\% on $vIoU@0.3$ for interrogative sentences between rows 3 and 4) over using time encoding only. 

\noindent \textbf{Initialization.}\label{sec:addinit}
We now provide detailed results for the ablation on our weight initialization.
The analysis is similar for both declarative and interrogative sentences.
In detail, Table~\ref{table:addinit} shows that pretraining is highly beneficial for spatio-temporal video grounding (+9.4\% on $vIoU@0.3$ for declarative sentences and +6.9\% on $vIoU@0.3$ for interrogative sentences between rows 1 and 3). 
The gain mainly comes from the spatial grounding performance (+11.7\% on $\siou{}$ for declarative sentences and +7.1\% on $\siou{}$ for interrogative sentences).
Additionally, we observe the benefit of using the spatial self-attention weights from the MDETR decoder to initialize the temporal self-attention in our decoder (+3.2\% on $vIoU@0.3$ for declarative sentences and +3.3\% on $vIoU@0.3$ for interrogative sentences between rows 2 and 3).

\begin{table*}[!htbp]
\vspace{-0pt}
\begin{center}
\setlength\tabcolsep{1.5pt}
\resizebox{1.\linewidth}{!}{
\begin{tabular}{lllll|ccccc|ccccc}
& \multirow{2}{*}{Slow} & \multirow{2}{*}{\makecell{\small{Spatial} \\ \small{Pool.}}}
& \multirow{2}{*}{f} & \multirow{2}{*}{g} & 
\multicolumn{5}{c}{Declarative Sentences} & \multicolumn{5}{c}{Interrogative Sentences} \\ 
& & & & & \small{m\_tIoU} & \small{m\_vIoU} & \small{vIoU@0.3} & \small{vIoU@0.5} & \small{\siou{}} & \small{m\_tIoU} & \small{m\_vIoU} & \small{vIoU@0.3} & \small{vIoU@0.5} & \small{\siou{}} \\
\hline
1. & \xmark & \xmark & Linear & Sum + Linear
& 42.7 & 18.6 & 25.0 & 14.8 & 39.6 & 
42.5 & 16.9 & 22.0 & 12.9 & 35.1 \\
2. & \cmark & - & 0 & 0 
& 46.2 & 24.9 & 34.4 & 21.8 & 49.7 & 
45.1 & 20.9 & 28.3 & 17.9 & 40.5 \\
3. & \cmark & \cmark & Linear & Sum + Linear 
& 45.8 & 25.0 & 34.7 & 22.1 & 50.2
& 44.9 & 21.1 & 29.2 & 17.8 & 40.9 \\
4. & \cmark & \xmark & Linear & Product + $\sigma$
& 46.2 & 26.2 & 36.0 & 23.9 & 52.0
& 45.4 & 22.1 & 30.1 & 18.8 & 43.0 \\  
5. & \cmark & \xmark & Transformer & Sum + Linear 
& \textbf{46.4} & 26.4 & \textbf{36.4} & 23.8 & \textbf{52.8}
& 45.3 & 22.2 & 30.2 & 19.6 & 43.3 \\  
6. & \cmark & \xmark & Linear & Sum + Linear 
& \textbf{46.4} & \textbf{26.6} & 36.1 & \textbf{24.7} & \textbf{52.8} 
& \textbf{45.6} & \textbf{22.5} & \textbf{30.8} & \textbf{19.8} & \textbf{43.6} \\ 
\end{tabular}}
\vspace{-0.3cm}
\caption{\small Comparison of designs for the video-text encoder, with or without the slow branch, with or without spatial pooling in the fast branch, with variants of the fast module $f$ and aggregation module $g$, on the VidSTG validation set.}
\label{table:highfre}
\vspace{-0.5cm}
\end{center}
\end{table*}

\noindent \textbf{Impact of spatial resolution and temporal stride $k$.}\label{sec:addresolution} 
In this section, we provide detailed results on the VidSTG dataset for the ablation on the impact of the spatial frame resolution and the temporal stride $k$.
The analysis is similar for both declarative and interrogative sentences.
In detail, Table~\ref{table:addresolution} shows that increasing the resolution is an important factor of performance for spatio-temporal video grounding (see rows 2 and 4).
Our proposed video-text encoder enables us to train on higher resolutions at a given memory usage.
This leads to a better performance-memory trade-off (rows 4, 5, 6, 8) compared to the baseline variant with temporal stride $k=1$ (row 1).
In particular, the best spatio-temporal video grounding results ($m\_vIoU$ and $vIoU@R$) are obtained with temporal stride $k=4$ and resolution 352 (row 6).

\noindent \textbf{Impact of the fast branch.} 
Finally, we provide detailed results on the VidSTG dataset for the ablation on the importance of our fast branch where we compare, for the best variant, temporal stride $k=4$ and resolution 352, our slow-fast video-text encoder to a slow-only variant.
The analysis is similar for both declarative and interrogative sentences.
By comparing rows 6 and 7 in Table~\ref{table:addresolution}, our fast branch significantly improves the spatio-temporal video grounding performance (+2.2\% $vIoU@0.3$ for declarative sentences and +1.9\% $vIoU@0.3$ for interrogative sentences) with low computational memory overhead.
This shows that the fast branch recovers useful spatio-temporal details lost by the temporal sampling operation in the slow branch.

\section{Additional Experiments}\label{sec:addexp}
In this section, we provide additional ablation studies. As in the ablations presented Section~\ref{sec:ablations}, unless stated otherwise, we use spatial frame resolution of 224 pixels and temporal stride $k=5$.

\noindent \textbf{Design of the fast and aggregation modules.}\label{sec:design}
Here we further ablate the fast and aggregation modules $f$ and $g$ used in our dual-branch encoder.
We report results in Table~\ref{table:highfre}.
The comparison between our slow-fast design (row 6) and the slow-only variant (row 2) is discussed in Section~\ref{sec:ablations}.
Likewise, we compare our slow-fast design to a fast-only variant (row 1). The fast-only variant does not use the slow multi-modal branch, in which case the video-text features are the fast visual-only features concatenated with the text features.
As shown in Table \ref{table:highfre}, our slow-fast design outperforms the fast-only variant, showing the importance of the slow multi-modal branch. 
We further compare the design of our fast and aggregation modules $f$ and $g$ (row 6) to other alternatives:
row 3, a variant with the same primitives $f$ and $g$ but with $f$ operating on features pooled over the spatial dimension;
row 4, a variant which uses the same fast module $f$ but a gating aggregation module $g(h_v(v, t), f(v)) = \sigma(h_v(v, t) * f(v))$ where $\sigma$ is the sigmoid function;
row 5, a variant that uses the same aggregation module $g$ but a fast temporal transformer module $f$, which models temporal interactions between spatially-detailed features.
As shown in Table \ref{table:highfre}, our design outperforms row 3, showing that preserving spatial information for each frame is crucial for the effectiveness of the fast branch. 
Additionally, our design slightly improves over row 4, indicating that further forcing the network to use the slow branch is not helpful.
Finally, our design slightly improves over row 5, suggesting that additional modeling of temporal interactions in our encoder is not necessarily helpful.

\section{Broader Impact}\label{sec:impact}
This work is a contribution to spatio-temporal video grounding and its potential positive or negative impacts depend on the application. 
Such models may be used for video surveillance and hence lead to questionable use. On the other hand, we believe that such methods could improve explainability of vision and language models which may help to understand some of their biases.  
This work also ablates memory usage when learning such models and thus could help promote development of lighter models with a reduced impact on the environment. 

\end{document}